\documentclass[11pt,letterpaper]{article}
\usepackage{color}
\usepackage[dvipsnames]{xcolor}
\usepackage{naaclhlt2016}
\usepackage{times}
\usepackage{latexsym}
\usepackage{graphicx}

\naaclfinalcopy % Uncomment this line for the final submission
\newcommand{\possign}{{\color{OliveGreen} $\bigtriangleup$}}
\newcommand{\negsign}{{\color{RedOrange} $\bigtriangledown$}}

\title{Sentiment Composition of Words with Opposing Polarities}

\author{Svetlana Kiritchenko \and Saif M. Mohammad\\
	    National Research Council Canada\\
	    {\tt \small \{svetlana.kiritchenko,saif.mohammad\}@nrc-cnrc.gc.ca}
}

\date{}

\begin{document}

\maketitle

\begin{abstract}
In this paper, we explore sentiment composition in phrases that have at least one positive and at least one negative word---phrases like {\it happy accident} and {\it best winter break}.  
We compiled a dataset of such opposing polarity phrases and manually annotated them with real-valued scores of sentiment association. 
Using this dataset, we analyze the linguistic patterns present in opposing polarity phrases. 
Finally, we apply several unsupervised and supervised techniques of sentiment composition to determine their efficacy on this dataset. 
Our best system, which incorporates information from the phrase's constituents, their parts of speech, their sentiment association scores, and their embedding vectors, obtains an accuracy of over 80\% on the opposing polarity phrases.
\end{abstract}

\section{Introduction}
\label{Introduction}

The Principle of Compositionality states that the meaning of an expression is determined by the meaning of its constituents and by its grammatical structure \cite{Montague1974}. 
By extension, sentiment composition is the determining of sentiment of a multi-word linguistic unit, such as a phrase or a sentence, based on its constituents. 
In this work, we study sentiment composition in phrases that include at least one positive and at least one negative word---for 
example, phrases such as \textit{happy accident}, \textit{couldn't stop smiling}, and \textit{lazy sundays}.
We refer to them as {\it opposing polarity phrases}. 
Such phrases present a particular challenge for automatic sentiment analysis systems that often rely on bag-of-word features.

Word--sentiment associations are commonly captured in {\it sentiment lexicons}. 
However, most existing manually created sentiment lexicons include only single words. 
Lexicons that include sentiment associations for {\it multi-word phrases} as well as their constituent words can be very useful in studying sentiment composition. 
We refer to them as {\it sentiment composition lexicons (SCLs)}.

We created a sentiment composition lexicon for opposing polarity phrases and their constituent words  
\cite{OPP-lrec}.\footnote{www.saifmohammad.com/WebPages/SCL.html\#OPP} 
Both phrases and single words were manually annotated with real-valued sentiment association scores using an annotation scheme known as Best--Worst Scaling.\footnote{ 
Best--Worst Scaling has been shown to produce reliable  real-valued sentiment association scores \cite{maxdiff-naacl2016}.}
We  refer to the created resource as the {\it Sentiment Composition Lexicon for Opposing Polarity Phrases (SCL-OPP)}.  
The lexicon includes entries for 265 trigrams, 311 bigrams, and 602 unigrams.

In this paper, we use SCL-OPP to analyze regularities present in different kinds of opposing polarity phrases. 
We calculate the extent to which  different part-of-speech combinations result in phrases of positive and negative polarity. 
We also show that for most phrases, knowing the parts of speech and polarities of their constituents is not enough to reliably predict the sentiment of the phrase.

We apply several unsupervised and supervised techniques of sentiment composition to determine their efficacy on predicting the sentiment of opposing polarity phrases. 
Our experiments indicate that the sentiment of the last unigram or the sentiment of the most polar unigram in the phrase are not strong predictors of the overall sentiment of the phrase. 
Similarly, adjectives and verbs do not always dominate the sentiment in such phrases. 
Finally, we show that the constituent words, their parts of speech, their sentiment association scores, and their embedding vectors are all useful features---a supervised sentiment composition system that incorporates them obtains accuracies over 80\% on both bigram and trigram opposing polarity phrases.

\section{Related Work}

A number of approaches have been proposed to address sentiment composition, which include manually derived syntactic rules 
\cite{moilanen2007sentiment,neviarouskaya2010recognition}, combination of hand-written rules and statistical learning \cite{choi2008learning}, and machine learning approaches \cite{nakagawa2010dependency,Yessenalina2011,dong2015statistical}. 
Much work has been devoted to model the impact of negators and (to a lesser degree) intensifiers, words commonly referred to as contextual valence shifters, on sentiment of words they modify \cite{Polanyi04,Kennedy05,liu2009review,Wiegand10,Taboada_2011,Kiritchenko2014}. 
\newcite{SCL-NMA} created a sentiment composition lexicon for negators, modals, and adverbs (SCL-NMA) through manual annotation and analyzed the effect of these groups of modifiers on sentiment in short phrases. 
Recently, recursive deep model approaches have been proposed for handling sentiment of syntactic phrases through sentiment composition over parse trees 
\cite{Socher2013,Zhu2014,irsoy2014deep,tai2015improved}.
In this work, we apply several unsupervised and supervised techniques of sentiment composition for a specific type of phrases---opposing polarity phrases. 

\section{Creating a Sentiment Lexicon for Opposing Polarity Phrases}
\label{lexicon-description}

\begin{table}[t!]
\begin{center}
\small{
\begin{tabular}{lr}
\hline {\bf Term} & {\bf Sentiment}\\
 & {\bf score}\\ \hline
best winter break &	0.844\\
breaking free	&  0.172\\
isn't long enough	& -0.188\\
breaking	& -0.500\\
heart breaking moment	& -0.797\\[5pt]
\hline
\end{tabular}
}
\caption{\label{tab:lex-examples} {Example entries in SCL-OPP.}
}
\end{center}
\vspace*{-3mm}
\end{table}

This section summarizes how we created a sentiment composition lexicon for opposing polarity phrases using the Best--Worst Scaling annotation technique. 
For more details we refer the reader to \cite{OPP-lrec}.  
Table~\ref{tab:lex-examples} shows a few example entries from the lexicon.

\noindent {\bf Term selection:} 
We polled the Twitter API (from 2013 to 2015) to collect about 11 million tweets that contain emoticons: `:)' or `:('. 
We will refer to this corpus as the {\it Emoticon Tweets Corpus}. 
From this corpus, we selected bigrams and trigrams that had at least one positive word and at least one negative word. 
The polarity labels (positive or negative) of the words were determined by simple look-up in  existing sentiment lexicons: Hu and Liu lexicon \cite{Hu04}, NRC Emotion lexicon \cite{MohammadT10,MohammadT13}, MPQA lexicon \cite{Wilson05}, and NRC's Twitter-specific lexicon \cite{Kiritchenko2014,MohammadSemEval2013}.\footnote{If a word was marked with conflicting polarity in two lexicons, then that word was not considered as positive or negative.}
In total, 576 opposing polarity $n$-grams (bigrams and trigrams) were selected. 
We also chose for annotation all unigrams that appeared in the selected set of bigrams and trigrams. 
There were 602 such unigrams. 
Note that even though the multi-word phrases and single-word terms were drawn from a corpus of tweets, most of the terms are used in everyday English.\\[-6pt]

\noindent {\bf Best--Worst Scaling Method of Annotation:} 
Best--Worst Scaling (BWS), also sometimes referred to as Maximum Difference Scaling (MaxDiff), is an annotation scheme that exploits the comparative approach to annotation \cite{Louviere_1990,Louviere2015}. 
Annotators are given four items (4-tuple) and asked which term is the Best (highest in terms of the property of interest) and which is the Worst (least in terms of the property of interest). 
Responses to the BWS questions can then be translated into real-valued scores
through a simple counting procedure: For
each term, its score is calculated as the percentage of times the term was chosen as the
Best minus the percentage of times the term was chosen as the Worst
\cite{Orme_2009}.
The scores range from -1 to 1. 

We employ Best--Worst Scaling for sentiment annotation by providing four (single-word or multi-word) terms at a time and asking which term is the most positive (or least negative) and which is the least positive (or most negative). 
Each question was answered by eight annotators through a crowdsourcing platform, CrowdFlower.\footnote{Let {\it majority answer} refer to the option most chosen for a question.
81\% of the responses matched the majority answer.} We refer to the resulting lexicon as the  {\it Sentiment Composition Lexicon for Opposing Polarity Phrases (SCL-OPP)}. 

Portions of the created lexicon have been used as development and evaluation sets in 
SemEval-2016 Task 7 `Determining Sentiment Intensity of English and Arabic Phrases' \cite{SemEval2016Task7}.\footnote{http://alt.qcri.org/semeval2016/task7/}  
The objective of that task was to test different methods of automatically predicting sentiment association scores for multi-word phrases.

\section{Sentiment Composition Patterns}
\label{patterns}

\setlength{\tabcolsep}{4pt}

\begin{table}[t]
\begin{center}
 \small{
\begin{tabular}{lrc}
\hline
  {\bf SCP}	&	{\bf Occ.} &{\bf \# phrases}\\
\hline
\negsign adj. + \possign adj. $\rightarrow$ \possign phrase	& 0.76 & 17\\
\negsign adj. + \possign noun $\rightarrow$ \negsign phrase	& 0.59 & 68\\
\possign adj. + \negsign noun $\rightarrow$ \negsign phrase	& 0.53 & 73\\
\possign adverb + \negsign adj. $\rightarrow$ \negsign phrase	& 0.89 & 18 \\
\possign adverb + \negsign verb $\rightarrow$ \negsign phrase	& 0.91 & 11\\
\negsign noun + \possign noun $\rightarrow$ \possign phrase	& 0.60 & 10\\
\possign noun + \negsign noun $\rightarrow$ \negsign phrase	& 0.52 & 25 \\
\negsign verb + det. + \possign noun $\rightarrow$ \negsign phrase	& 0.65 & 17\\
\negsign verb + \possign noun $\rightarrow$ \negsign phrase	& 0.82 & 17\\
\hline
\end{tabular}
}
\caption{Sentiment composition patterns (SCPs) in SCL-OPP. \possign denotes a positive word or phrase, \negsign denotes a negative word or phrase. `Occ.' stands for occurrence rate of an SCP.}
\label{tab:patterns}
\end{center}
\vspace*{-2mm}
\end{table}

SCL-OPP allows us to explore sentiment composition patterns in opposing polarity phrases. 
We define a {\it Sentiment Composition Pattern (SCP)} as a rule that includes on the left-hand side the parts of speech (POS) and the sentiment associations of the constituent unigrams (in the order they appear in the phrase), and on the right-hand side the sentiment association of the phrase. Table~\ref{tab:patterns} shows examples.
SCPs that have a positive phrase on the right-hand side will be called {\it positive SCPs}, whereas
SCPs that have a negative phrase on the right-hand side will be called {\it negative SCPs}. 
 Below are some questions regarding SCPs and opposing polarity phrases that we explore here: 
\begin{itemize}
\vspace*{-1mm}
\item Which SCPs are common among opposing polarity phrases?
\vspace*{-1mm}
\item With the same left-hand side of an SCP, how often is the composed phrase positive and how often is the composed phrase negative? For example, when negative adjectives combine with a positive noun, how often is the combined phrase negative?
\vspace*{-1mm}
\item Are some parts of speech (of constituent words) more influential in determining the sentiment of a phrase than others? 
\end{itemize}
\vspace*{-1mm}
To answer these questions, each of the entries in SCL-OPP 
is marked with the appropriate SCP. The part-of-speech sequence of a phrase is determined by looking up the most common part-of-speech sequence for that phrase in the Emoticon Tweets Corpus.\footnote{The corpus was automatically POS tagged using the CMU Tweet NLP tool \cite{Gimpel11}.}
Next, for every left-hand side of an SCP, we determine the ratio of `how often occurrences of such combinations in SCL-OPP resulted in a positive phrase' to `how often such combinations were seen in total'. We will refer to these scores as the {\it occurrence rates (`Occ.') of positive SCPs}. The {\it occurrence rates of negative SCPs} are calculated in a similar manner.

Table~\ref{tab:patterns} presents all SCPs with the left-hand side combination appearing at least ten times in SCL-OPP, and whose occurrence rate is equal to or greater than 50\%.
For example, the second row tells us that there are 68 bigrams in SCL-OPP such that  the first word is a negative adjective and the second word is a positive noun. 
Out of these 68 bigrams, 59\% are negative, and the remaining 41\% are positive, so the occurrence rate of this pattern is 0.59.

The most common SCPs in our lexicon are ``\possign adj. + \negsign noun $\rightarrow$ \negsign phrase'' (73) and ``\negsign adj. + \possign noun $\rightarrow$ \negsign phrase'' (68).
Observe that the occurrence rates of the patterns are spread over the entire range from 52\% to 91\%. 
Only two patterns have very high occurrence rates (around 90\%): ``\possign adverb + \negsign adj. $\rightarrow$ \negsign phrase'' and ``\possign adverb + \negsign verb $\rightarrow$ \negsign phrase''.
Thus, for most opposing polarity phrases, their sentiment cannot be accurately determined based on the POS and sentiment of the constituents alone.

Both SCPs with high occurrence rates include adverbs that serve as intensifiers---words that increase or decrease the degree of association of the following word with positive (negative) sentiment (e.g., \textit{incredibly slow}, \textit{dearly missed}). 
Only the degree of association for the next word is changed while its polarity (positive or negative) is often preserved. 
Some adjectives can also play the role of an intensifier when combined with another adjective (e.g., \textit{crazy talented}) or a noun (e.g., \textit{epic fail}). 
For example, the adjective \textit{great}, often considered highly positive, becomes an intensifier when combined with some nouns (e.g., \textit{great loss}, \textit{great capture}). 
Other adjectives determine the polarity of the entire phrase (e.g., \textit{happy tears}, \textit{bad luck}). 
Therefore, the occurrence rates of patterns like ``\negsign adj. + \possign noun $\rightarrow$ \negsign phrase'' are low. 
Overall, even though adjectives and verbs are frequently the primary source of sentiment in the phrase, some nouns can override their sentiment as in \textit{new crisis} or \textit{leave a smile}. 
SCL-OPP includes phrases corresponding to many different kinds of sentiment composition patterns, and therefore, it is a useful resource for studying linguistic underpinnings of sentiment composition as well as for evaluating sentiment composition algorithms for opposing polarity phrases.  

\section{Automatically Predicting Sentiment}
\label{cv-classification}

We now investigate whether accurate models of sentiment composition for opposing polarity phrases can be learned. 
We conduct experiments with several baseline unsupervised classifiers as well supervised techniques using features, such as unigrams, POS, sentiment scores, and word embeddings. 

The problem of sentiment composition can be formulated in two different ways:  
a binary classification task where the system has to predict if the phrase is positive or negative;  
and a regression task where the system has to predict the real-valued sentiment association score of the phrase. 
We evaluate binary classification with simple accuracy ({\it acc.}) and the regression task with Pearson correlation coefficient ($r$). 
Learning and evaluation are performed separately for bigrams and trigrams.

\subsection{Baseline Classifiers}
\label{unsupervised}

The oracle `majority label' baseline assigns to all instances the most frequent polarity label in the dataset. 
The `last unigram' baseline returns the sentiment score (or the polarity label) of the last unigram in the phrase. 
For the regression task, we use the real-valued sentiment score of the unigram whereas for the binary classification task we use the polarity label (positive or negative). 
The `most polar unigram' baseline assigns to the phrase the sentiment score (or the polarity label) of the most polar word in that phrase, i.e., the word with the highest absolute sentiment score.
The `part-of-speech (POS) rule' baseline assigns sentiment as follows: 
\vspace*{-4mm} 
\begin{enumerate}
	\item If the phrase has an adjective, return the sentiment score (polarity) of the last adjective;
\vspace*{-3mm} 
	\item Else, if the phrase has a verb, return the sentiment score (polarity) of the last verb;
\vspace*{-3mm} 
	\item Else, return the sentiment score (polarity) of the most polar word.
\end{enumerate}
\vspace*{-3mm}

\subsection{Supervised Classifiers}
\label{supervised}

We train a Support Vector Machines classifier with RBF kernel for the binary classification task and a Support Vector regression model with RBF kernel for the regression task using the LibSVM package \cite{CC01a}. 
For both tasks, the models are trained using different combinations of the following features obtained from the target phrase: all unigrams, POS tag of each unigram, sentiment label of each unigram, sentiment score of each unigram, and the word embedding vector for each unigram. 
The word embeddings are obtained by running  word2vec software \cite{mikolov2013efficient} on the Emoticon Tweets Corpus. 
We use the skip-gram model with the default parameters and generate 200-dimensional vectors for each unigram present in the corpus. 
For each task, ten-fold cross-validation is repeated ten times, and the results are averaged. 

\subsection{Results}
\label{results}

\setlength{\tabcolsep}{3pt}

\begin{table}[t]
\begin{center}
\resizebox{0.49\textwidth}{!}{
\begin{tabular}{llcccc}
\hline
  & &	\multicolumn{2}{c}{\bf Binary} & \multicolumn{2}{c}{\bf Regression}\\
  \multicolumn{2}{l}{\bf Features}	&	\multicolumn{2}{c}{\bf (Acc.)} & \multicolumn{2}{c}{\bf (Pearson $r$)}\\
	& & {2-gr} & {3-gr} & {2-gr} & { 3-gr}\\
\hline
\multicolumn{2}{l}{\bf Baselines}\\
a. & majority label & 56.6 & 60.8 & - & - \\
b. & last unigram & 57.2 & 59.3 & 0.394 & 0.376\\
c. & most polar unigram & 66.9 & 69.8 & 0.416 & 0.551\\
d. & POS rule & 65.6 & 63.8 & 0.531 & 0.515\\[2pt]
\multicolumn{2}{l}{\bf Supervised classifiers}\\
e. & POS + sent. label & 65.7 & 64.2 & - & -\\
f. & POS + sent. score & 74.9 & 74.8 & 0.662 & 0.578\\
g. & row f + uni & 82.0 & 81.3 & 0.764 & 0.711\\
h. & row f + emb(avg) + emb(max) & 78.2 & 79.5 & 0.763 & 0.710\\
i. & row f + emb(conc) & 80.2 & 76.5 & 0.790 & 0.719\\
j. & row f + emb(conc) + uni &\bf 82.6 &\bf 80.9 & \bf 0.802 &\bf  0.753\\
k. & POS + emb(conc) + uni & 76.3 & 80.2 & 0.735 & 0.744\\
\hline
\end{tabular}
}
\caption{Performance of the automatic systems on SCL-OPP. Features used: unigrams (uni), part-of-speech of a unigram (POS), sentiment binary label of a unigram (sent. label), sentiment real-valued score of a unigram (sent. score), embeddings (emb). `emb(conc)' is the concatenation of the embedding vectors of the constituent unigrams; 
`emb(avg)' is the average vector of the unigram embeddings; `emb(max)' is maximal vector of the unigram embeddings.
}
\label{tab:res-cv}
\end{center}
\vspace*{-3mm}
\end{table}

The results for all baseline and supervised methods are presented in Table~\ref{tab:res-cv}.
The `majority label', `last unigram', `most polar unigram', and `POS rule' baselines are shown in rows a to d.
Observe that the sentiment association of the last unigram is not very predictive of the phrase's sentiment (row b).\footnote{Note that the results for the `last unigram' baseline are still better than the results of random guessing (acc = 50, $r$ = 0). For the majority of $n$-grams in SCL-OPP, the polarity of the first unigram is opposite to the polarity of the last unigram. Thus, the results for a similar `first unigram' baseline (not shown here) are worse than those obtained by the `last unigram' baseline.} 
Both the `most polar unigram' and the `POS rule' classifiers perform markedly better than the majority baseline. 
Interestingly, the `most polar unigram' classifier outperforms the slightly more sophisticated `POS rule' approach on most tasks. 
Also, we found that within bigram phrases that contain adjectives or verbs, the adjective or verb constituents are the most polar words in only about half of the instances (and even less so in trigrams).  This indicates that adjectives and verbs do not always dominate the sentiment in a phrase.

The results obtained using supervised techniques with various feature combinations are presented in rows e to k (Table~\ref{tab:res-cv}).  
Using only POS and binary sentiment labels of the constituent unigrams, the supervised learning algorithm does not perform much better than our `POS rule' baseline   
(the accuracies in row e are just slightly higher than those in row d). 
With access to real-valued sentiment scores of unigrams much more accurate models can be learned (row f). 
Furthermore, the results show that the sentiment of a phrase depends on its constituent words and not only on the sentiment of the constituents (row g shows markedly better performance than row f; all the differences are statistically significant, $p < .01$).
Concatenating word embeddings was found to be more effective than averaging. (Averaging is common when creating features for sentences). 
Having access to both unigrams and word embedding features produces the best results. (The differences between the scores in row i and row j are statistically significant, $p < .01$.) 
Row k shows results of the model trained without the gold sentiment scores of the unigrams. 
Observe that for bigrams, there is a substantial drop in performance compared to row j (6.3-point drop in accuracy on the binary task, 6.7-point drop in Pearson correlation on the regression task) whereas for trigrams the performance is not affected as much (less than 1-point change on both tasks). 
Thus, having access to sentiment scores of constituents is particularly useful for determining sentiment of bigram phrases.

\section{Conclusions}

We created a real-valued sentiment composition lexicon for opposing polarity phrases and their constituent words, through manual annotation.  
We analyzed patterns of sentiment composition across phrases formed with different POS combinations. 
Further, we applied several unsupervised and supervised techniques of sentiment composition to determine their efficacy on opposing polarity phrases. 
We showed that for most phrases the sentiment of the phrase cannot be reliably predicted only from the parts of speech and sentiment association of their constituent words, and that the constituent words, their parts of speech, their sentiment scores, and their embedding vectors are all useful features in supervised sentiment prediction on this dataset.

We intend to use SCL-OPP in the following applications: 
(1) to automatically create a large coverage sentiment lexicon of multi-word phrases and apply it in downstream applications such as sentence-level sentiment classification, and (2) to investigate how the human brain processes sentiment composition. 

\bibliography{maxdiff}
\bibliographystyle{naaclhlt2016}

\end{document}